\documentclass[10pt,twocolumn,letterpaper]{article}

\usepackage[pagenumbers]{cvpr} 

\usepackage{graphicx}
\usepackage{amsmath}
\usepackage{amssymb}
\usepackage{booktabs}
\usepackage{epsfig}
\usepackage{algorithm}
\usepackage{algorithmic}
\usepackage{float}
\usepackage{lipsum}
\newfloat{figtab}{htb}{fgtb}
\makeatletter
  \newcommand\figcaption{\def\@captype{figure}\caption}
  \newcommand\tabcaption{\def\@captype{table}\caption}
  \newcommand\algcaption{\def\@captype{algorithm}\caption}
\makeatother
\usepackage{booktabs}
\usepackage{amsfonts}
\usepackage{tabularx}
\usepackage{animate}
\usepackage{multirow}
\usepackage{pifont}
\usepackage[accsupp]{axessibility}  
\usepackage{appendix}

\usepackage[pagebackref,breaklinks,colorlinks]{hyperref}

\usepackage[capitalize]{cleveref}
\crefname{section}{Sec.}{Secs.}
\Crefname{section}{Section}{Sections}
\Crefname{table}{Table}{Tables}
\crefname{table}{Tab.}{Tabs.}

\def\cvprPaperID{6704} 
\def\confName{CVPR}
\def\confYear{2023}

\begin{document}

\title{MOSO: Decomposing MOtion, Scene and Object for Video Prediction}

\author{
Mingzhen Sun $^{1,2}$ \hspace{7mm} Weining Wang $^1$ \hspace{7mm} Xinxin Zhu $^1$ \hspace{7mm} Jing Liu $^{1,2,*}$\\
$^1$The Laboratory of Cognition and Decision Intelligence for Complex Systems,\\
Institute of Automation, Chinese Academy of Sciences (CASIA)\\
$^2$School of Artificial Intelligence, University of Chinese Academy of Sciences (UCAS)\\
{\tt\small sunmingzhen2020@ia.ac.cn \{weining.wang, xinxin.zhu, jliu\}@nlpr.ia.ac.cn}
}

\maketitle
\newcommand\blfootnote[1]{%
\begingroup
\renewcommand\thefootnote{}\footnote{#1}%
\addtocounter{footnote}{-1}%
\endgroup
}

\begin{abstract}
 \blfootnote{* Corresponding Author}
Motion, scene and object are three primary visual components of a video. In particular, objects represent the foreground, scenes represent the background, and motion traces their dynamics. Based on this insight, we propose a  two-stage MOtion, Scene and Object decomposition framework (MOSO)\footnote{Codes have been released in \textit{https://github.com/iva-mzsun/MOSO}} for video prediction, consisting of MOSO-VQVAE and MOSO-Transformer. In the first stage, MOSO-VQVAE decomposes a previous video clip into the motion, scene and object components, and represents them as distinct groups of discrete tokens. Then, in the second stage, MOSO-Transformer predicts the object and scene tokens of the subsequent video clip based on the previous tokens and adds dynamic motion at the token level to the generated object and scene tokens. Our framework can be easily extended to unconditional video generation and video frame interpolation tasks. Experimental results demonstrate that our method achieves new state-of-the-art performance on five challenging benchmarks for video prediction and unconditional video generation: BAIR, RoboNet, KTH, KITTI and UCF101. In addition, MOSO can produce realistic videos by combining objects and scenes from different videos.
\end{abstract}

\section{Introduction}
Video prediction aims to generate future video frames based on a past video without any additional annotations \cite{vp_intro1,vp_intro2}, which is important for video perception systems, such as autonomous driving \cite{vp_autonomous}, robotic navigation \cite{vp_robot} and decision making in daily life \cite{vp_decision}, etc.  
Considering that video is a spatio-temporal record of moving objects, an ideal solution of video prediction should depict visual content in the spatial domain accurately and predict motions in the temporal domain reasonably. 
However, easily distorted object identities and infinite possibilities of motion trajectories make video prediction a challenging task.

\begin{figure}
    \centering
    \includegraphics[width=1\linewidth]{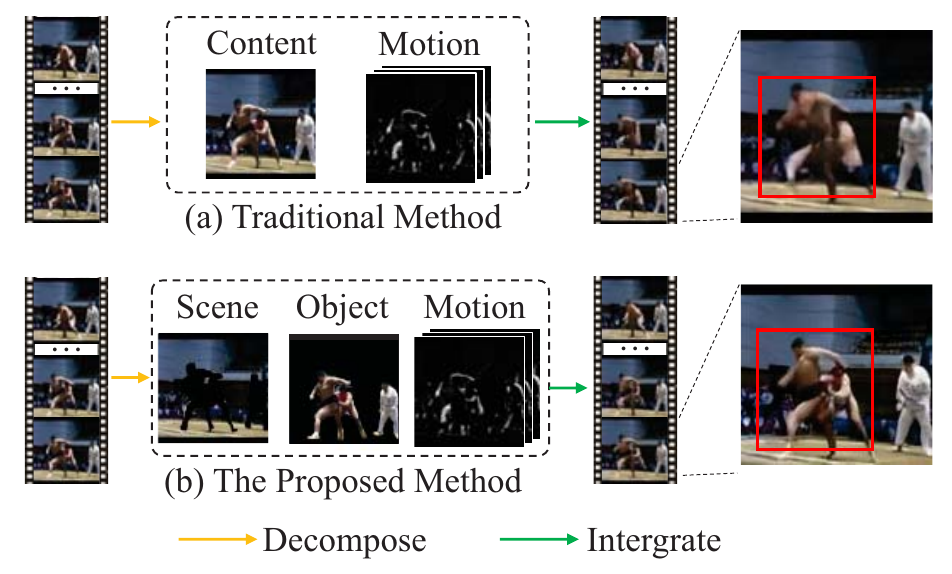}
    \vspace{-4mm}
    \caption{
    Rebuilding video signals based on (a) traditional decomposed content and motion signals or (b) our decomposed scene, object and motion signals.
    Decomposing content and motion signals causes blurred and distorted appearance of the wrestling man, while further separating objects from scenes resolves this issue.
    }
    \vspace{-2mm}
    \label{fig:intro}
\end{figure}

Recently, several works \cite{mocovp,mocovp2} propose to decompose video signals into content and motion, with content encoding the static parts, i.e., scene and object identities, and motion encoding the dynamic parts, i.e., visual changes.
This decomposition allows two specific encoders to be developed, one for storing static content signals and the other for simulating dynamic motion signals.
However, these methods do not distinguish between foreground objects and background scenes, which usually have distinct motion patterns.
Motions of scenes can be caused by camera movements or environment changes, e.g., a breeze, whereas motions of objects such as jogging are always more local and routine.
When scenes and objects are treated as a unity, 
their motion patterns cannot be handled in a distinct manner, resulting in blurry and distorted visual appearances.
As depicted in Fig. \ref{fig:intro}, it is obvious that the moving subject (i.e., the wrestling man) is more clear in the video obtained by separating objects from scenes than that by treating them as a single entity traditionally.

Based on the above insight, we propose a two-stage MOtion, Scene and Object decomposition framework (MOSO) for video prediction. 
We distinguish objects from scenes and utilize motion signals to guide their integration. 
In the first stage, MOSO-VQVAE is developed to learn motion, scene and object decomposition encoding and video decoding in a self-supervised manner. 
Each decomposed component is equipped with an independent encoder to learn its features and to produce a distinct group of discrete tokens.
To deal with different motion patterns, we integrate the object and scene features under the guidance of the corresponding motion feature.
Then the video details can be decoded and rebuilt from the merged features.
In particular, the decoding process is devised to be time-independent, so that a decomposed component or a single video frame can be decoded for flexible visualization.

In the second stage, MOSO-Transformer is proposed to generate a subsequent video clip based on a previous video clip.
Motivated by the production of animation, which first determines character identities and then portrays a series of actions, MOSO-Transformer firstly predicts the object and scene tokens of the subsequent video clip from those of the previous video clip.
Then the motion tokens of the subsequent video clip are generated based on the predicted scene and object tokens and the motion tokens of the previous video clip.
The predicted object, scene, and motion tokens can be decoded to the subsequent video clip using MOSO-VQVAE.
By modeling video prediction at the token level, MOSO-Transformer is relieved from the burden of modeling millions of pixels and can instead focus on capturing global context relationships.
In addition, our framework can be easily extended to other video generation tasks, 
including unconditional video generation and video frame interpolation tasks, by simply revising the training or generation pipelines of MOSO-Transformer.

Our contributions are summarized as follows:

$\bullet$ We propose a novel two-stage framework MOSO for video prediction, which could decompose videos into motion, scene and object components and conduct video prediction at the token level.

$\bullet$ MOSO-VQVAE is proposed to learn motion, scene and object decomposition encoding and time-independently video decoding in a self-supervised manner, which allows video manipulation and flexible video decoding. 

$\bullet$ MOSO-Transformer is proposed to first determine the scene and object identities of subsequent video clips and then predict subsequent motions at the token level.

$\bullet$ Qualitative and quantitative experiments on five challenging benchmarks of video prediction and unconditional video generation demonstrate that our proposed method achieves new state-of-the-art performance.

\section{Related Work}
\textbf{Video Prediction}
The video prediction task has received increasing interest in the computer vision field.
ConvLSTM \cite{convlstm} combines CNN and LSTM architectures and adopts an adversarial loss.
MCnet \cite{mocovp} models pixel-level future video prediction with motion and content decomposition for the first time.
GVSD \cite{gvsd} proposes a spatio-temporal CNN combined with adversarial training to untangle foreground objects from background scenes, while severe distortion of object appearances exists in their predicted video frames.
MCVD \cite{mcvd} adopts a denoising diffusion model to conduct several video-related tasks conditioned on past and/or future frames.
Although previous models can predict consistent subsequent videos, they still suffer from indistinct or distorted visual appearances since they lack a stable generator or fail to decouple different motion patterns.
SLAMP \cite{slamp} and vid2vid \cite{vid2vid} decomposes video appearance and motion for video prediction with the help of optical flow.
SADM \cite{sadm} proposes a semantic-aware dynamic model that predicts and fuses the semantic maps (content) and optical flow maps (motion) of future video frames.
In addition to optical flow and semantic maps, Wu et al. \cite{wu2020future} further utilizes instance maps to help separate objects from backgrounds.
Although these works also decompose video components, they are more complicated than MOSO since they require much more additional information.
Furthermore, these previous works are primarily based on generative adversarial networks or recurrent neural networks, while MOSO follows a recently developed two-stage autoregressive generation framework, which demonstrates greater potential on open domain visual generation tasks.

\textbf{Two-stage Visual Generation}
The two-stage framework is first proposed for image generation \cite{dalle,cogview,taming} and demonstrates excellent generation ability.
Motivated by the success, several attempts have been made to extend the two-stage framework to video generation tasks \cite{godiva,videogpt,taming,videogpt}.
For video prediction, MaskViT \cite{maskvit} encodes videos by frame though VQ-GAN \cite{taming} and models video tokens with a bidirectional Transformer through window attention.
For unconditional video generation, VideoGPT \cite{videogpt} encodes videos by employing 3D convolutions and axial attention, and then models video tokens in an auto-regressive manner.
However, existing two-stage works for video tasks do not consider video component decomposition and are affected by flicker artifacts and expensive computation costs.

\section{MOSO}
\begin{figure*}
    \centering
    \includegraphics[width=1\linewidth]{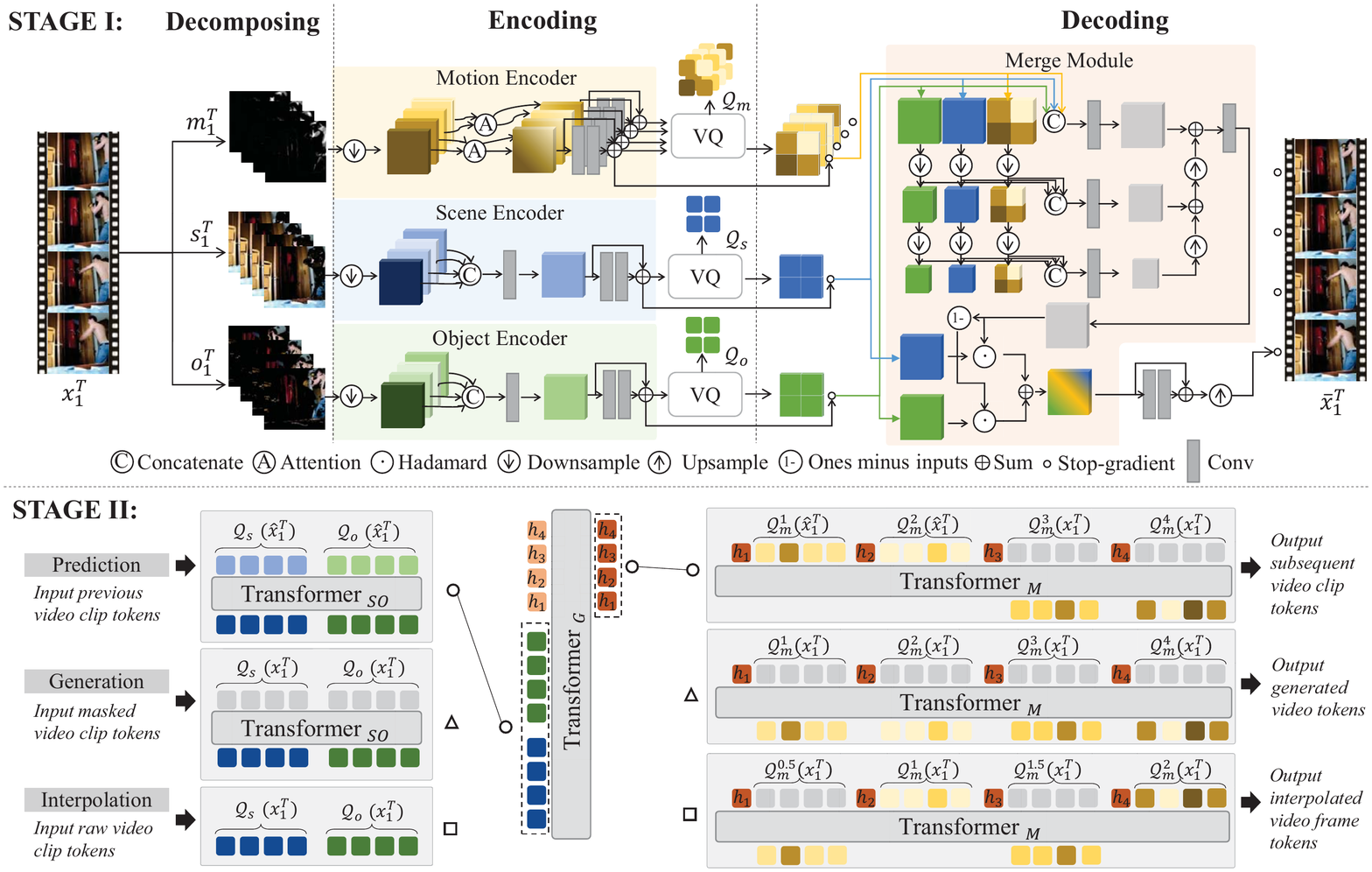}
    \vspace{-6mm}
    \caption{
    The overall framework of MOSO.
    Above the dashed line shows the architecture of MOSO-VQVAE, which decomposes and encodes the motion, scene and object components of a video, produces token groups and reconstructs the input video from features.
    Below the dashed line shows the generation process of MOSO-Transformer, where the small gray grid represents the mask token.
    For video prediction, Transformer$_{SO}$ generates the scene and object tokens of $x_1^T$ based on those of the pseudo video $\hat x_1^T$, which is composed of the given video frames.
    Then Transformer$_{G}$ outputs the guidance embeddings $\{h_1,...,h_T\}$ and Transformer$_{M}$ predicts the subsequent motion tokens, where $K$ is 2, $T$ is 4 and the generation iteration $S$ is 1 for illustration.
    MOSO-Transformer can perform unconditional video generation by feeding all mask tokens and video frame interpolation by generating only motion tokens of interpolated video frames.
    }
    \label{fig:moso}
    \vspace{-1mm}
\end{figure*}

In this section, we present our proposed framework MOSO in detail.
MOSO is a novel two-stage framework for video prediction and consists of MOSO-VQVAE and MOSO-Transformer, where MOSO-VQVAE encodes decomposed video components to tokens and MOSO-Transformer conducts video prediction at the token level.
The overall framework of MOSO is drawn in Fig. \ref{fig:moso}.

We denote a $T$-frame video as $x_1^T$, which is an abbreviation for $\{x_t\}_{t=1}^T$, where $x_t \in R^{H \times W \times C}$ is the $t$-th video frame, $H$ is the height, $W$ is the weight and $C$ is the number of channels.
Video prediction requires predicting a subsequent video clip $x_{K+1}^{T}$ based on a previous one $x_{1}^{K}$.

\subsection{Stage I: MOSO-VQVAE}
MOSO-VQVAE consists of a codebook $\mathbb{E}$, three video encoders (i.e., motion, scene and object encoders) and a time-independent video decoder.
The codebook is denoted as  $\mathbb{E} = \{e_i\}_{i=1}^N$, where $e_i\in R^D$ is the $i$-th codebook entry, $N$ is the codebook size and  $D$ is the dimension.
MOSO-VQVAE first roughly decomposes the motion, scene and object components of the input video $x_1^T$ with a preprocessing algorithm, obtaining the motion video $m_1^T$, the scene video $s_1^T$ and the object video $o_1^T$.
Then the video encoder of each component takes its corresponding video as input and learns its feature.
Each feature is then quantized using the codebook and a distinct group of tokens is available as a by-product.
Finally, the video decoder integrates the quantized features to reconstruct the input video.

\vspace{-2mm}
\subsubsection{Decomposing}
Considering that most of the scene motion is caused by camera movements or environment changes, 
scene motion is typically either tiny or extremely dramatic.
Based on this observation, we propose an efficient preprocessing algorithm to separate one video into three distinct videos (i.e., motion, scene and object videos) without additional annotations. 
Specifically, frame difference is calculated and used as the motion video $m_1^T$.
Then a lower threshold $c_{lb}$ and an upper threshold $c_{ub}$ are adopted to filter pixels with middle differences to obtain the object video $o_1^T$.
The scene video $s_1^T$ is composed of the left pixels with differences less than $c_{lb}$ or higher than $c_{ub}$, corresponding to tiny or dramatic movements.
More details are presented in the appendix.

\subsubsection{Encoding}
The scene encoder consists of several downsample layers, a temporal compressing layer and a residual stack.
Given a scene video, the downsample layers, i.e., 2D convolutions with stride 2, downsample each frame by a factor of $f$.
Then the frame features are concatenated in the channel dimension and compose feature $z'_{s} \in R^{H/f \times W/f \times TD}$, where $TD$ is the number of channels.
The temporal compressing layer, i.e., linear projection $R^{TD \to D}$, reduces the number of channels to $D$.
The residual stack, composed of several residual layers \cite{resnet}, then learns the scene feature $z_{s}$.
The object encoder has the same structure as the scene encoder but takes an object video as input and outputs an object feature $z_{o}$.

The motion encoder replaces the temporal compressing layer in the scene encoder with a temporal self-attention layer and outputs motion feature $z_{m}$.
In particular, the downsampled frame features are concatenated in the temporal dimension and split into $N_t$ parts, composing feature 
$z'_{m} \in R^{H/f \times W/f \times N_t \times (T / N_t) \times D}$.
Then the temporal self-attention is conducted on each $Z \in \{{z'_m}^{h,w,n}\}$:
\begin{equation}
\begin{aligned}
    Q =& Z W_Q,  K=Z W_K,  V=Z W_V \\
    Y &= softmax(QK^{\top} / \sqrt{D})V
    \label{eq:attention}
\end{aligned}
\end{equation}
where $h \in \{1,...,\frac{H}{f}\}, w \in \{1,...,\frac{W}{f}\}, n\in\{1,...,N_t\}$ and $W_Q$, $W_K$ and $W_V$ are learnable $D \times D$ projection matrices. 

The codebook is used to quantize each feature $z \in \{z_s, z_o, z_m\}$ and obtain discrete tokens as:
\begin{gather}
\begin{aligned}
\mathcal{VQ}&(x_1^T)^{h,w,(t)} = e_r, \quad \mathcal{Q}(x_1^T)^{h,w,(t)} = r \\
r &= \mathop{\arg\min}_{1 \leq i \leq N} \lVert e_i - z^{h,w,(t)} \rVert_2^2 \\
\end{aligned}
\end{gather}
where $h \in \{1,...,\frac{H}{f}\}$, $w \in \{1,...,\frac{W}{f}\}$, $t \in \{1,...,T\}$,
$z_{s}, z_{o} \in R^{H/f \times W/f \times D}$, $z_{m} \in R^{H/f \times W/f \times T \times D}$,
$\mathcal{VQ}$ obtains quantized features,  $\mathcal{Q}$ obtains maps of tokens,
$e_i$ is the $i$-th codebook entry and $\lVert * \rVert^2$ denotes the calculation of L2 norm.
The quantized features have the same shape as the raw features.
Note that the temporal dimension of the scene and object features is 1, which is omitted for conciseness.

\subsubsection{Decoding}
The video decoder consists of a merge module, a residual stack and several upsample layers, i.e., transposed 2D convolutions with stride 2.
To allow gradients to back-propagate to the encoders, the video decoder stops gradients of each quantized feature, obtaining $\Tilde{z} = sg(\mathcal{VQ}(x_1^T) - z) + z$, where $sg$ denotes the operator of stop-gradient.

The merge module dynamically integrates the object feature $\Tilde{z}_{o}$ and the scene feature $\Tilde{z}_{s}$ according to the $t$-th motion feature $\Tilde{z}_{m}^t$ to obtain the video feature $z^t$ for the $t$-th video frame.
It first obtains three multi-scale weight features:
\begin{gather}
w^t_1 = \mathcal{F}_1([\Tilde{z}_{o}, \Tilde{z}_{s}, \Tilde{z}_{m}^t]) \\
w^t_2 = \mathcal{F}_2([\downarrow_2 (\Tilde{z}_{o}), \downarrow_2 (\Tilde{z}_{s}), \downarrow_2 (\Tilde{z}_{m}^t)]) \\
w^t_3 = \mathcal{F}_3([\downarrow_4 (\Tilde{z}_{o}), \downarrow_4 (\Tilde{z}_{s}), \downarrow_4 (\Tilde{z}_{m}^t)])
\end{gather}
where $[*]$ concatenates input features along the channel dimension and $\mathcal{F}_1$, $\mathcal{F}_2$ and $\mathcal{F}_3$ are linear functions $R^{3D \to D}$.
$\downarrow_f$ denotes a downsample function implemented by 2D convolutions with stride 2, where $f$ denotes the sample factor.
The weight features are then merged by:
\begin{gather}
w^t_4 = w^t_2 + \uparrow_{2}(w^t_3) \\
w^t = \sigma( \mathcal{F}( w^t_1 + \uparrow_{2}(w^t_4) ) )
\end{gather}
where $\sigma(*)$ denotes the sigmoid activation function and $\mathcal{F}: R^{D} \to R^D$ is a linear function.
$\uparrow_f$ denotes an upsampling function implemented by bilinear interpolation, where $f$ denotes the sample factor.
The $t$-th video frame feature is then obtained by:
\begin{gather}
z^t = \Tilde{z}_{s} \odot w^t + \Tilde{z}_{o} \odot (1 - w^t)
\end{gather}
where $\odot$ denotes Hadamard product.

The residual stack has the same structure as that in the encoders.
It operates on video frame features and is blind to their temporal index.
Finally, several upsample layers spatially upsample video features to reconstruct video frames.
The reconstructed $t$-th video frame is denoted as $\bar x_t$.

Notably, our video decoder is time-independent since the merge module, the residual stack and the upsample layers are agnostic to the temporal position of each video frame and have no access to other motion features $\Tilde{z}_{m}^k, k \neq t$ when reconstructing the $t$-th video frame.
In other words, we only need to obtain the scene, object and $t$-th motion features when decoding the $t$-th video frame.
It not only allows flexible visualization but also facilitates the modeling process in the second stage.
In particular, the decomposed scenes and objects can be decoded and visualized by the video decoder based on their tokens, as discussed in the appendix.

\begin{figure*}
    \centering
    \includegraphics[width=0.95\linewidth]{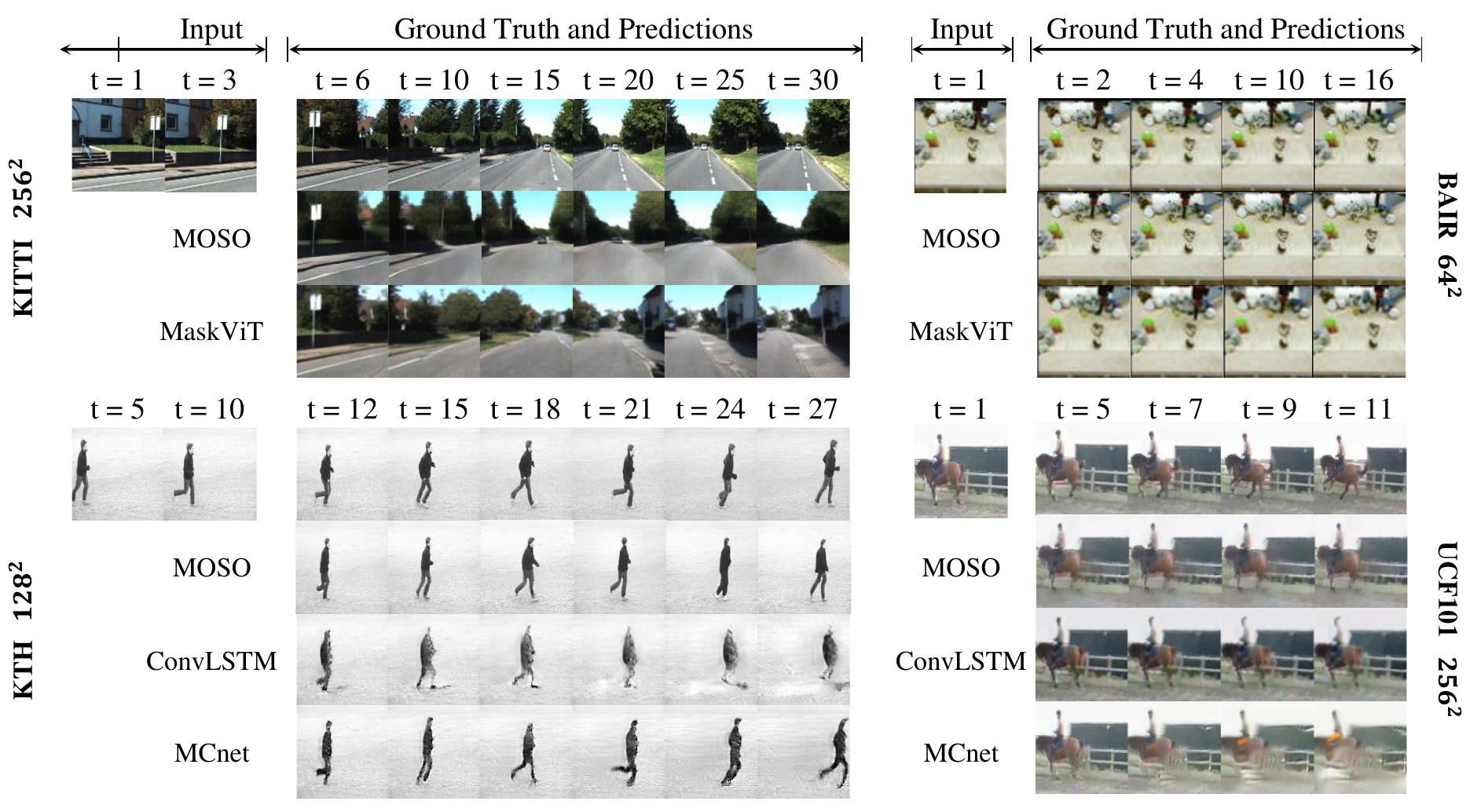}
    \vspace{-2mm}
    \caption{Qualitative comparison of MOSO with MaskViT \cite{maskvit}, ConvLSTM \cite{convlstm} and MCnet \cite{mocovp} on test set of KITTI $256^2$, BAIR $64^2$, KTH $128^2$ and UCF101 $256^2$ for video prediction.
    }
    \label{fig:samples_vp}
    \vspace{-4mm}
\end{figure*}
\subsubsection{Training}
The overall loss function $\mathcal{L}$ for training MOSO-VQVAE consists of the reconstruction loss $\mathcal{L}_{rec}$ and the commitment loss $\mathcal{L}_{com}$.
The reconstruction loss $\mathcal{L}_{rec}$ optimizes the video encoders and the video decoder by minimizing the L2 distance and the perceptual loss \cite{lpips} between each pair of input and reconstructed video frames.
The commitment loss $\mathcal{L}_{com}$ optimizes the encoders to ensure the training of the encoders keeps in step with that of the codebook, thus preventing divergence:
\begin{equation}
    \begin{aligned}
    \mathcal{L}_{com} = \sum_{h=1}^{H/f}  \sum_{w=1}^{W/f} & \bigg [ 
    \sum_{t=1}^{T} \Big\lVert z^{h,w,(t)} - sg\Big(\mathcal{VQ}(x_1^T)^{h,w,(t)}\Big) \Big\rVert^2 \bigg ]
    \end{aligned}
\end{equation}
where $sg$ denotes the operator of stop-gradient and $\lVert * \rVert^2$ denotes the calculation of L2 norm.
The exponential moving average (EMA) optimization method is used to optimize the codebook entries as in \cite{vqvae2}. 

\subsection{Stage II: MOSO-Transformer}
MOSO-Transformer is comprised of three bidirectional Transformers: Transformer$_{SO}$, Transformer$_{G}$ and Transformer$_{M}$.
Given a previous video clip $x_1^K$, MOSO-Transformer predicts the subsequent video clip $x_{K+1}^T$ at the token level though pretrained MOSO-VQVAE.
Since MOSO-VQVAE takes fixed length $T$-frame videos as input, we first pad the previous video clip $x_1^K$ to $T$ frames by duplicating the last frame, obtaining a pseudo video $\hat x_1^T$:
\begin{equation}
    \hat x_1^{T}=\{x_1,...,x_{K-1},x_K,\underbrace{x_K,...,x_K}_{T-K}\}, \quad K \leq T
    \label{eq:hatx}
    \vspace{-2mm}
\end{equation}
Then the motion, scene and object tokens of $\hat x_1^T$ can be produced by MOSO-VQVAE.
Based on the scene and object tokens of $\hat x_1^{T}$, Transformer$_{SO}$ is trained to generate the scene and object tokens of $x_1^{T}$:
\begin{equation}
    P(\mathcal{Q}_{s}(x_1^{T}), \mathcal{Q}_{o}(x_1^{T}); \mathcal{Q}_{s}(\hat x_1^{T}), \mathcal{Q}_{o}(\hat x_1^{T}))
    \label{eq:t1}
\end{equation}
To provide guidance for the generation of motion tokens, Transformer$_{G}$ is used to learn the guidance embeddings $\{h_1,...,h_T\}$ based on the scene and object tokens of $x_1^{T}$.
Considering that both Transformer$_{SO}$ and Transformer$_{G}$ require learning the scene and object content at the token level, we train them with shared parameters.

Rethinking the temporal split operation demonstrated in Eq. (\ref{eq:attention}), 
when $T$ can be exactly divided by $K$ and the temporal dimension is $N_t=\frac{T}{K}$ parts, 
the motion tokens of the target video $x_1^T$ and the pseudo video $\hat x_1^T$ satisfy:
\begin{equation}
    \mathcal{Q}(x_1^T)^t == \mathcal{Q}(\hat x_1^T)^t, \quad 1 \leq t \leq K
\end{equation}
Namely, the motion tokens for the first $K$ video frames in the target and pseudo videos are totally the same.
A detailed explanation of this property and a more general situation are presented in the appendix.

Based on the above property, 
Transformer$_{M}$ only needs to generate motion tokens of the last $T-K$ frames of $x_1^T$ based on the guidance embeddings $\{h_1,...,h_T\}$ and the motion tokens of the first $K$ given video frames of $\hat x_1^T$.
Following the training pipeline in \cite{maskgit}, we first randomly mask $\gamma(r)$ proportion of motion tokens with mask tokens $[M]$, 
where $r$ is a scalar and $\gamma(r)$ is a monotonically declining function with $\gamma(0)=1$ and $\gamma(1)=0$.
Then Transformer$_{M}$ is trained through a cross-entropy loss to model the distribution of masked motion tokens based on unmasked motion tokens, the guidance embeddings and the motion tokens of the first $K$ video frames of $\hat x_1^T$:
\begin{equation}
\begin{aligned}
    P(\{\mathcal{Q}_{m}(x_1^{T})&^n_{=[M]}\}; \{\mathcal{Q}_{m}(x_1^{T})^n_{\not = [M]}\}, \{\mathcal{Q}_{m}(\hat x_1^{T})^r\}, \{h_t\}) \\
     K+1 \leq &n \leq T, \quad 1 \leq r \leq K, \quad 1 \leq t \leq T
    \label{eq:t3}
\end{aligned}
\end{equation}

\begin{table*}
    \begin{minipage}[htp]{0.38\linewidth}
    \centering
    \caption{
        Quantitative comparison with other methods on BAIR for video prediction.
    }
    \label{tab:bair}
    \scalebox{0.8}{
    \begin{tabularx}{1.27\linewidth}{ p{0.5\linewidth}| X<{\centering} | X<{\centering}}
        \toprule
        \textbf{Method}             &   Params  &   FVD $\downarrow$ \\
        \midrule
        LVT \cite{lvt}              &   -       &   125.8   \\
        SAVP \cite{savp}            &   -       &   116.4   \\
        DVD-GAN-FP \cite{dvd}       &   -       &   109.8   \\
        VT (S) \cite{vt}            &   46M    &   106.0    \\
        TrIVD-GAN-FP \cite{trivd}   &   -       &   103.3   \\
        VideoGPT \cite{videogpt}    &   -       &   103.3   \\
        CCVS \cite{ccvs}            &   -       &   99.0      \\
        VT (L) \cite{vt}            &   373M    &   94.0    \\
        MaskViT \cite{maskvit}      &   189M    &   93.7    \\
        FitVid \cite{fitvid}        &   302M    &   93.6    \\
        MCVD \cite{mcvd}            &   251M    &   89.5   \\
        RaMViD \cite{ramvid}        &   -       &   89.2    \\
        NUWA \cite{nuwa}            &   870M    &   86.9    \\
        \midrule
        MOSO                        &   265M    &\textbf{83.6}  \\
        \bottomrule
    \end{tabularx}
    }
  \end{minipage}
  \begin{minipage}[htp]{0.60\linewidth}
    \begin{minipage}[htp]{1\linewidth}
        \centering
        \caption{Comparison with other methods on RoboNet for video prediction.}
        \vspace{-3mm}
        \label{tab:robonet}
        \scalebox{0.8}{
        \small
        \begin{tabularx}{1.1\linewidth}{ p{0.2\linewidth}| X<{\centering} | X<{\centering} | X<{\centering} X<{\centering} X<{\centering} X<{\centering}}
            \toprule
            \textbf{Method}         &   Params  &$H\times W$&   FVD $\downarrow$    &  PSNR $\uparrow$  &   SSIM $\uparrow$ &   LPIPS $\downarrow$    \\
            \midrule
    MaskViT \cite{maskvit}  &   257M    &64$^2$ &   133.5               &   23.2            &   80.5            &   0.042   \\
    SVG \cite{svg}          &   298M    &64$^2$ &   123.2               &   23.9            &   87.8            &   0.060   \\
    GHVAE \cite{ghvae}      &   599M    &64$^2$ &   95.2                &   24.7            &   89.1            &   0.036   \\
    FitVid  \cite{fitvid}   &   302M    &64$^2$ &   62.5                &   28.2            &   89.3            &   0.024   \\
    MOSO                    &   265M    &64$^2$ &   \textbf{53.3}       &  \textbf{31.2}    &   \textbf{92.1}   &   \textbf{0.017}  \\
            \midrule
    MaskViT \cite{maskvit}  &   228M    &256$^2$ &   211.7               &   20.4            &   67.1            &   0.170   \\
    MOSO                    &   265M    &256$^2$ &   \textbf{91.5}       &  \textbf{26.3}    &   \textbf{79.9}   &   \textbf{0.096}  \\
            \bottomrule
        \end{tabularx}}
    \end{minipage}
    \begin{minipage}[htp]{1\linewidth}
        \centering
        \caption{Comparison with other methods on KITTI for video prediction.}
        \vspace{-3mm}
        \label{tab:kitti}
        \scalebox{0.8}{
        \small
        \begin{tabularx}{1.1\linewidth}{ p{0.2\linewidth}| X<{\centering} | X<{\centering} | X<{\centering} X<{\centering} X<{\centering} X<{\centering}}
            \toprule
            \textbf{Method} &   Params  & $H\times W$ &   FVD $\downarrow$    &  PSNR $\uparrow$  &   SSIM $\uparrow$ &   LPIPS $\downarrow$    \\
            \midrule
    SVG \cite{svg}          &   298M    &   $64^2$  &   1217.3          &   15.0            &   41.9            &   0.327   \\
    FitVid  \cite{fitvid}   &   302M    &   $64^2$  &   884.5           &   17.1            &   49.1            &   0.217   \\
    GHVAE \cite{ghvae}      &   599M    &   $64^2$  &   552.9           &   15.8            &   51.2            &   0.286   \\
    MaskViT \cite{maskvit}  &   181M    &   $64^2$  &   401.9           &   \textbf{27.2}   &   58.1            &   0.089   \\
    MOSO                    &   177M    &   $64^2$  &   \textbf{395.3}  &  25.6             &   \textbf{74.8}   &   \textbf{0.086}  \\
            \midrule
    MaskViT \cite{maskvit}  &   228M    &   $256^2$  &  \textbf{446.1}  &  \textbf{26.2}    &   40.7            &   0.270   \\
    MOSO                    &   219M    &   $256^2$  &  516.6           &  21.1             &   \textbf{59.2}   &   \textbf{0.265}  \\
            \bottomrule
        \end{tabularx}
        }
    \end{minipage}
  \end{minipage}
\end{table*}

\begin{table}[t]
    \centering
    \caption{
    Comparison with prior works on KTH for video prediction.
    }
    \vspace{-3mm}
    \label{tab:kth}
    \scalebox{0.8}{
    \begin{tabularx}{1.2\linewidth}{ p{0.35\linewidth}| X<{\centering} | X<{\centering} X<{\centering} X<{\centering}}
        \toprule
        \textbf{Method}         &$H\times W$&  PSNR $\uparrow$  &   SSIM $\uparrow$ &   LPIPS $\downarrow$    \\
        \midrule
        SVG-LP \cite{svg-lp}    & 64$^2$    &   23.9           &   80.0           &   0.129   \\
        Struct-VRNN \cite{vrnn} & 64$^2$    &   24.3           &   76.6           &   0.124   \\
        SV2P \cite{sv2p}        & 64$^2$    &   25.9           &   78.2           &   0.232   \\
        SAVP \cite{savp}        & 64$^2$    &   26.0           &   80.6           &   0.116   \\
        MCVD \cite{mcvd}        & 64$^2$    &   26.4            &   81.2            &   -       \\
        GK \cite{grid}          & 64$^2$    &   27.1           & \textbf{83.7}      &   0.092   \\
        \midrule
        MOSO                    & 64$^2$    &  \textbf{29.8}    &   82.2   &   \textbf{0.083}  \\
        \bottomrule
    \end{tabularx}}
    \vspace{-3mm}
\end{table}

\vspace{-4mm}
\subsubsection{Predicting}
Inspired by the production of animation, which first creates figure characters and then depicts a series of actions, 
we predict the subsequent video clip by first determining its scene and object identities and then producing dynamic motions at the token level.
In particular, Transformer$_{SO}$ firstly outputs the distribution of the scene and object tokens of $x_1^T$ as in Eq. (\ref{eq:t1}).
Then the scene and object tokens are randomly sampled and the guidance embeddings can be obtained through the Transformer$_G$.
Given the guidance embeddings and a template filled with mask tokens $[M]$, Transformer$_{M}$ performs the following two steps for $S$ iterations to predict the motion tokens:
(1) outputting distribution of motion tokens as specified in Eq. (\ref{eq:t3}) and sampling motion tokens in the masked positions;
(2) randomly masking $\gamma(s/S)$ proportion of all motion tokens with previous unmasked tokens fixed, where $s$ denotes the $s$-th iteration.
Finally, the predicted video frames $x_{K+1}^T$ are decoded from the generated tokens by MOSO-VQVAE.
In addition, our MOSO-Transformer can be easily extended to other tasks involving unconditional video generation and video frame interpolation as shown in Fig. \ref{fig:moso}.

\section{Experiments}
\label{sec:exp}
In this section, we compare our method with prior works for video prediction and unconditional video generation tasks on five challenging benchmarks, including BAIR \cite{bair}, RoboNet \cite{robonet}, KTH \cite{kth}, KITTI \cite{kitti} and UCF101 \cite{ucf}.
Fig. \ref{fig:samples_vp} shows the qualitative comparison of video prediction.
We start with introducing evaluation metrics and experimental implementations.

\textbf{Metrics}
We adopt five evaluation metrics: Fr$\acute{e}$chet Video Distance (FVD), Fr$\acute{e}$chet Instance Distance (FID), Peak Signal-to-Noise Ratio (PSNR), Structural Similarity Index Measure (SSIM) and Learned Perceptual Image Patch Similarity (LPIPS).
FVD \cite{fvd} measures the distance between distributions of predicted videos and real-world videos.
We calculate FVD with codes released by StyleGAN-V\footnote{https://github.com/universome/stylegan-v} \cite{stylegan-v}, which has been proven to be a precise implementation of the official one.
To obtain the FVD score, we pass those given previous video frames and calculate between the predicted frames and corresponding ground truths following MaskViT \cite{maskvit}.
For unconditional video generation, we follow StyleGAN-V \cite{stylegan-v} and calculate FVD on 2048 randomly sampled videos.
FID \cite{fid} evaluates the generated videos by frame.
PSNR \cite{psnr}, SSIM \cite{ssim} and LPIPS \cite{lpips} measure frame-wise similarity between predicted and ground truth videos. 
We conduct one trial per video on the BAIR datasets.
On the KTH, KITTI and RoboNet datasets, we follow \cite{maskvit,fitvid} and report the best SSIM, PSNR and LPIPS scores over 100 trials per video to account for the stochastic nature of video prediction.
FVD is calculated over all 100 trials with batch size being 256 following \cite{maskvit}.

\textbf{Implementation}
MOSO-VQVAE encodes videos by clip.
When training MOSO-VQVAE, each video clip has a fixed length $T$, which is set as 16 for BAIR, 12 for RoboNet and 20 for KTH and KITTI.
$N_t$ is set as 1 for BAIR, 6 for RoboNet, 2 for KTH and 4 for KITTI.
For UCF101, $T$ and $N_t$ are set as 12 and 3 for video prediction or 16 and 1 for unconditional video generation.
The batch size is 32 for $64^2$ resolution and 4 for others.
The codebook size $N$ is 16384.
The preprocessing algorithm is used for the first 50k iterations.
After that, the scene and object encoders take raw video frames as input and learn scene and object identities in a self-supervised manner, and the motion encoder still takes the frame difference as input.
For MOSO-Transformer, the batch size is 16 for $64^2$ resolution and 4 for others.
$gamma(*)$ is set to decay as cosine and $S$ is set as 16.
For long video prediction, tokens of the last video clip are used as previous tokens for iterative prediction.
Deepspeed \cite{deepspeed} and mixed-precision FP16/FP32 \cite{mixed} are utilized for fast training.
Experiments are conducted on 4 A100s. 

\subsection{Video Prediction}
\textbf{BAIR.} The BAIR robot pushing dataset \cite{bair} records random movements of robotic arms.
Following prior work \cite{nuwa}, we predict subsequent 15 video frames given only 1 previous video frame, and all videos are resized to $64 \times 64$ resolution.
Qualitative comparison of MOSO with MaskViT is given in Fig. \ref{fig:samples_vp}.
Our MOSO achieves new state-of-the-art performance on this dataset as reported in Table \ref{tab:bair}, outperforming prior best work by 3.3 FVD.

\textbf{RoboNet.} The RoboNet dataset \cite{robonet} contains more than 15 million videos of 7 robotic arms pushing things in different bins.
Following prior works \cite{fitvid,fitvid108}, we randomly select 256 videos for testing and predict 10 subsequent video frames given 2 previous video frames.
As shown in Table \ref{tab:robonet}, MOSO achieves significant improvements on all metrics than prior works at both $64^2$ and $256^2$ resolution.

Both MOSO and MaskViT follow the pipeline of VQVAE and Transformer, while MOSO outperforms MaskViT by a large margin on the RoboNet dataset.
We attribute this phenomenon to two reasons.
Firstly, MaskViT models videos by frame, which can lead to flicker artifacts in videos with static backgrounds like RoboNet \cite{maskvit}, while MOSO models videos by clip and thus obtains better content consistency in videos.
Secondly, videos in RoboNet are easy to distinguish between foreground objects and background scenes, which facilitates object and scene decomposition and hence favors MOSO.
Quantitative results of MOSO-VQVAE video reconstruction are given in the appendix, which demonstrate that MOSO-VQVAE can handle videos with static and tiny-motion backgrounds quite well.


\textbf{KTH.} The KTH dataset \cite{kth} contains videos of 25 people performing 6 types of actions.
Following \cite{grid,mocovp}, we adopt videos of persons 1-16 for training and 17-25 for testing.
When predicting videos, 10 subsequent video frames are generated based on 10 previous video frames during training, and 40 subsequent video frames are required to be predicted during testing.
Following \cite{mocovp}, we manually trim the conditional video clips to ensure humans are always present.
We report the quantitative results in Table \ref{tab:kth} and the qualitative results in Fig. \ref{fig:samples_vp}.
More results are available in the appendix.
We do not calculate FVD since its batch size setting is ambiguously stated in \cite{grid}, which may dramatically impact the evaluation result.
As reported in Table \ref{tab:kth}, MOSO outperforms GK \cite{grid} by 0.27 on PSNR and 0.09 on LPIPS, and obtains a comparable SSIM score.
As depicted in Fig. \ref{fig:samples_vp}, our MOSO outperforms the previous motion and content decomposition method MCnet \cite{mocovp} with more distinct object appearances and less artifacts.

\textbf{KITTI.}
The KITTI dataset \cite{kitti} is a challenging dataset with only 57 training videos and dramatic moving scenes.
MaskViT \cite{maskvit} outperforms prior works by a large margin on this dataset, and we achieve comparable performance with MaskViT at both $64^2$ and $128^2$ resolutions as reported in Table \ref{tab:kitti}.
Different from the VQ-GAN adopted in MaskViT, which sets the weight of the adversarial loss as 1, MOSO-VQVAE adopts a smaller loss weight 0.1 as specified in the ablation study.
Despite such loss weight helps MOSO-VQVAE achieve outstanding performance on other datasets, it seems to be too small for KITTI to compete with the perceptual loss with weight 1, leading to some checkboard artifacts in the predicted video frames as shown in Fig. \ref{fig:samples_vp}.
These checkboard artifacts are produced by the ResNet-50 network used in the perceptual loss as indicated in \cite{peco,artifact}.



\subsection{Other Video Generation Tasks}
\textbf{Unconditional Video Generation.}
MOSO-Transformer can be trained for unconditional video generation by replacing the scene and object tokens of the given video clip with mask tokens and removing given motion tokens.
We quantitatively compare MOSO with other models on the UCF101 dataset for unconditional video generation as reported in Table \ref{tab:ucf}.
The results demonstrate that our MOSO outperforms the previous method \cite{stylegan-v} by 219.1 on FVD.

\begin{table}[t]
    \centering
        \caption{Quantitative comparison with other methods on UCF101 for unconditional video generation.}
        \vspace{-3mm}
        \label{tab:ucf}
        \scalebox{0.8}{
        \begin{tabularx}{1.2\linewidth}{ p{0.7\linewidth}| X<{\centering}}
            \toprule
            \textbf{Method}                 &  FVD $\downarrow$    \\
            \midrule
            MoCoGAN \cite{mocogan}          &   2886.9      \\
            +StyleGAN2 backbone             &   1821.4      \\
            MoCoGAN-HD \cite{mocogan-hd}    &   1729.6      \\        
            VideoGPT \cite{videogpt}        &   2880.6      \\
            DIGAN \cite{digan}              &   1630.2      \\
            StyleGAN-V \cite{stylegan-v}    &   1431.0      \\
            \midrule
            MOSO                            &  \textbf{1202.6}  \\
            \bottomrule
        \end{tabularx}}
        \vspace{-2mm}
\end{table}

\begin{figure}
    \centering
    \includegraphics[width=1\linewidth]{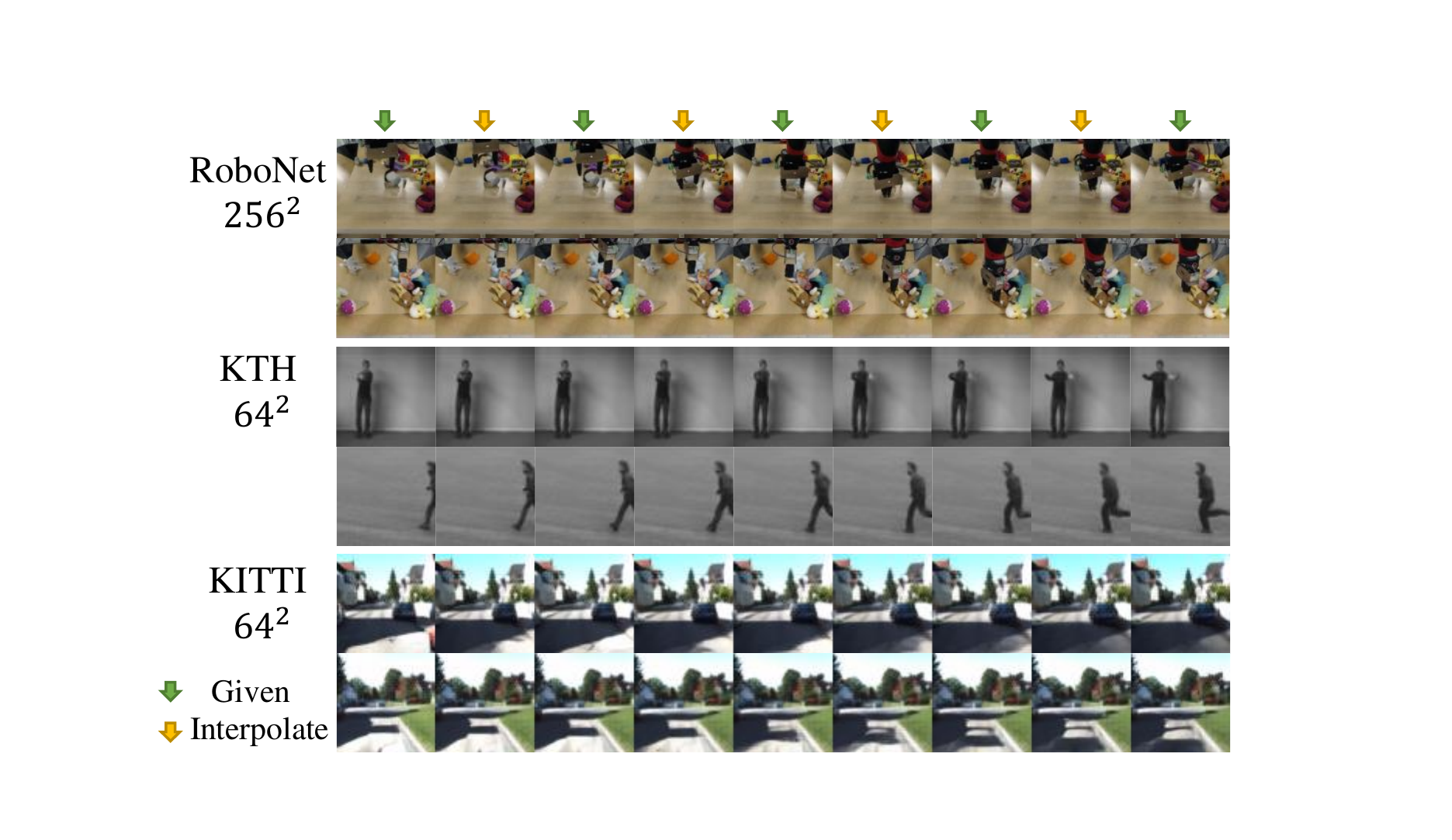}
    \caption{Samples of video frame interpolation on RoboNet, KTH and KITTI.}
    \label{fig:interpolate}
    \vspace{-4mm}
\end{figure}

\textbf{Video Frame Interpolation.}
MOSO can directly perform video frame interpolation after being trained for video prediction.
Based on raw motion, scene and object tokens, MOSO-Transformer generates interpolated video frames by initializing their motion tokens with mask tokens and then removing mask tokens through $S$ steps, where $S$ is 16.
Samples on the RoboNet, KTH and KITTI datasets are given in Fig. \ref{fig:interpolate}, which shows that MOSO could interpolate consistent video frames.

\begin{figure}
    \centering
    \includegraphics[width=0.95\linewidth]{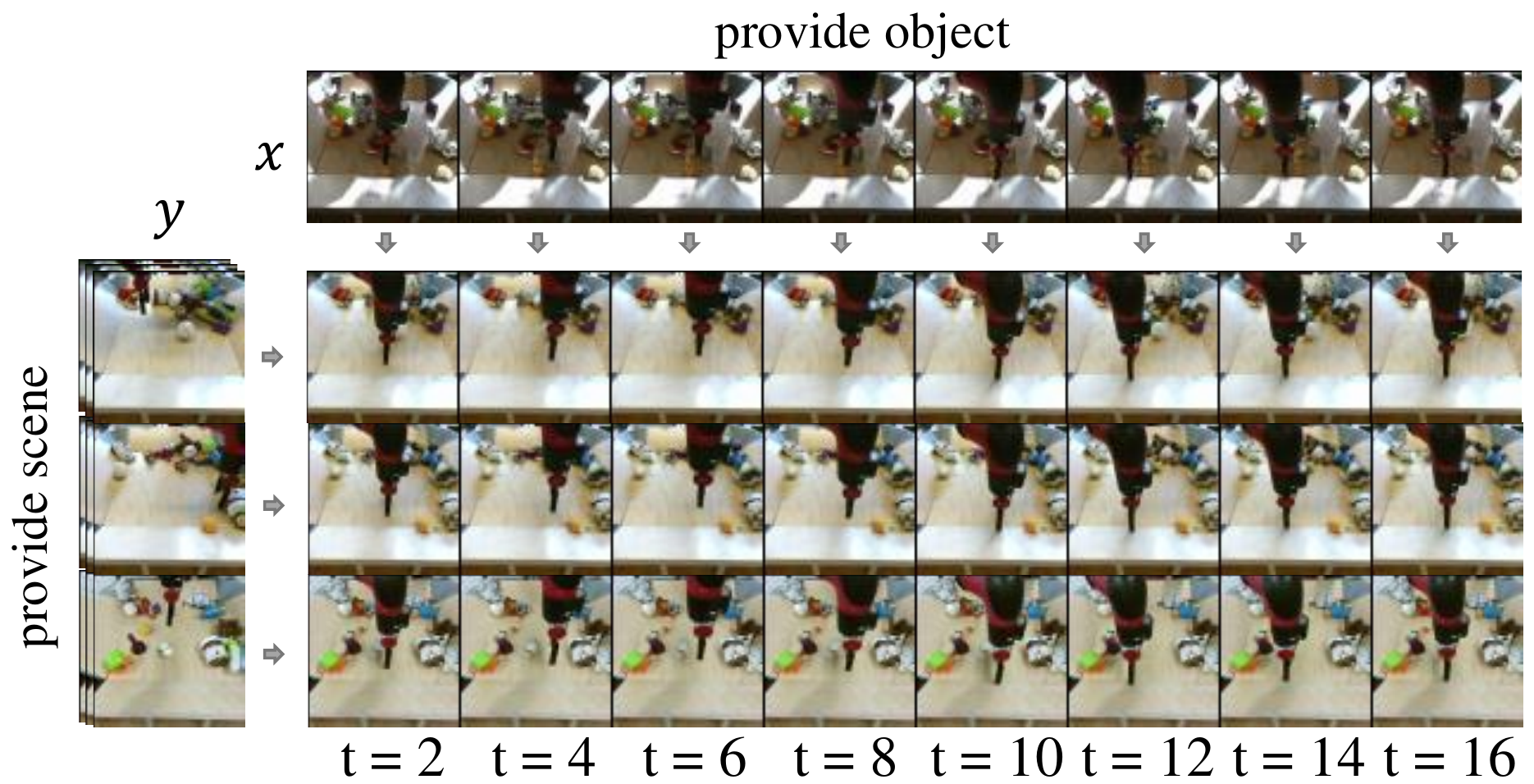}
    \caption{Samples of video manipulation on BAIR.
    Videos with objects from $x$ and scenes from $y$ are created by decoding object and motion tokens of $x$ combined with scene tokens of $y$.
    }
    \vspace{-3mm}
    \label{fig:swap}
\end{figure}

\textbf{Video Manipulation.}
By separating object identities from scenes, MOSO-VQVAE can decode realistic videos with objects from one video and scenes from other videos.
In particular, given two series of video frames $x_1^T$ and $y_1^T$, motion, scene and object tokens are obtained through the video encoders.
By combining the object and motion tokens of $x_1^T$ and the scene tokens of $y_1^T$, 
a new video with objects from $x_1^T$ and scene from $y_1^T$ can be decoded by the video decoder as shown in Fig. \ref{fig:swap}.

\subsection{Ablation Study}
We conduct ablation studies on the KTH $64^2$ dataset for video prediction and report the results in Table \ref{tab:ablate}.
The number of tokens for the non-decomposition or content and motion decomposition encoders are shown in Table \ref{tab:token}.
For a fair comparison, we ensure the parameters of both VQVAE and the Transformer of all involved models are comparable.
As reported in Table \ref{tab:ablate}, the performance of MOSO outperforms the non-decomposition method and the motion and content decomposition method by a large margin.

\begin{table}[t]
    \centering
    \caption{Ablation studies on KTH $64^2$ for video prediction.
    \textit{non decom.}: non-decomposition.
    \textit{mo. co.}: motion and content decomposition.
    \textit{mo. sc. ob.}: motion, scene and object decomposition.
    \textit{pre. alg.} denotes the preprocessing algorithm.
     \textit{Trans.} denotes Transformer.
    }
    \vspace{-3mm}
    \label{tab:ablate}
    \scalebox{0.8}{
    \begin{tabularx}{1.2\linewidth}{ p{0.4\linewidth} | p{0.25\linewidth}<{\centering} | X<{\centering} X<{\centering} X<{\centering}}
    \toprule
         Method             & Parameters (VAE+Trans.)  & PSNR$\uparrow$  & SSIM$\uparrow$  & LPIPS$\downarrow$     \\
         \midrule
         non decom. ($f_{non}:8$)         & 605M+308M  & 21.5  & 62.8  & 0.202  \\
         \midrule
         mo. co. ($f_{mo}$:8,$f_{co}$:4)            & 611M+265M  & 25.4  & 68.8  & 0.147 \\
         mo. co. ($f_{mo}$:8,$f_{co}$:2)            & 535M+265M  & 27.1  & 75.5  & 0.129 \\
         \midrule
         mo. sc. ob. (MOSO) & 593M+265M  & \textbf{29.8}  & \textbf{82.2}  & \textbf{0.083} \\
         MOSO - pre. alg.  & 593M+265M  & 29.2  & 82.0  & 0.086 \\
         MOSO - merge Module & 557M+265M  & 28.3  & 79.3  & 0.125 \\
         MOSO + single Trans. & 593M+286M  & 28.9  & 80.1  & 0.116 \\
    \bottomrule
    \end{tabularx}
    }
\end{table}

\begin{table}[t]
    \centering
    \caption{Number of tokens changes with downsample factors.}
    \vspace{-3mm}
    \label{tab:token}
    \scalebox{0.8}{
    \begin{tabularx}{1.2\linewidth}{ p{0.23\linewidth} | X<{\centering} X<{\centering}| X<{\centering} X<{\centering} | p{0.2\linewidth}<{\centering} }
    \toprule
    Method          & \multicolumn{2}{c|}{$f_{non}$}    & \multicolumn{2}{c|}{($f_m$,$f_c$)}    & ($f_m$,$f_s$,$f_o$)  \\
    \midrule
    Down. factor    & 8     & 4     & (8,4)    & (8,2)    & (8,4,4)     \\
    Token length    & 1280  & 5120  & 1536     & 2304     & 1792   \\
    FLOPs($\times10^{9}$)& 389   & 2382  & 481      & 796      & 581     \\
    \bottomrule
    \end{tabularx}
    }
    \vspace{-4mm}
\end{table}

\section{Conclusion and Discussions}
In this paper, we propose a novel two-stage motion, scene and object decomposition framework for video prediction.
Extensive experiments show that our method achieves new state-of-the-art performance on several challenging benchmarks, 
demonstrating the importance of decomposing the motion, scene and object video components.

Our research has two main limitations.
First, although the preprocessing algorithm is efficient and effective, it is not delicate enough.
More expressive and powerful tools, e.g. optical flow, may help decompose the motion, scene and object components better.
Second, we do not enlarge the model and dataset for video prediction, while several works \cite{nuwa,dalle} have revealed the potential of a huge Transformer for the open-domain visual generation.
Future works are highly-encouraged for further exploration.

\section{Acknowledgments}
We thank Yonghua Pan, Jiawei Liu and Zihan Qin for reviewing early drafts and helpful suggestions.
This work was supported by the National Key Research and Development Program of China (No. 2020AAA0106400) and National Natural Science Foundation of China (61922086, 61872366, U21B2043, 62102419, 62102416), and was sponsored by CAAI-Huawei MindSpore Open Fund.

\appendix
\section{Preprocessing Algorithm}
\label{sec:preprocess}
\begin{figure}[t]
    \centering
    \includegraphics[width=1\linewidth]{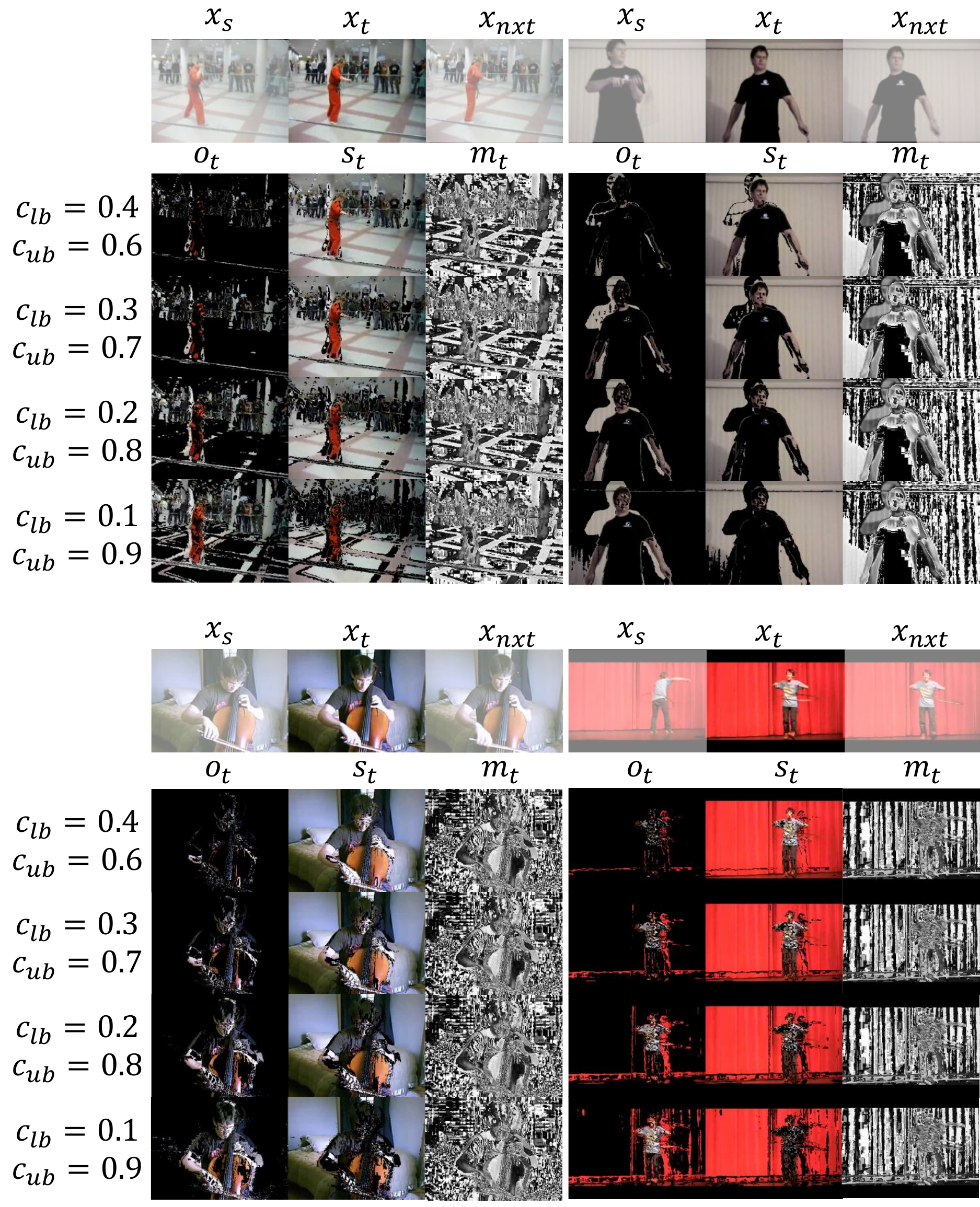}
    \caption{
    Visualizing the $t$-th frame in the decomposed motion video $m_1^T$, scene video $s_1^T$ and object video $o_1^T$ respectively through the preprocessing algorithm with different $c_{lb}$ and $c_{ub}$.
    }
    \label{fig:preprocess}
\end{figure}

We propose an efficient preprocessing algorithm for decomposing a video into motion, scene and object videos.
The pseudo-code for the preprocessing algorithm is presented in Algorithm. \ref{alg:frame}.
In particular, frame difference is calculated and employed as the motion video $m_1^T$.
Then, a lower threshold $c_{lb}$ and an upper threshold $c_{ub}$ are set to filter pixels with modest differences to obtain the object video $o_1^T$.
Finally, the left pixels are used to compose the scene video $s_1^T$.
In Fig. \ref{fig:preprocess}, we show decomposed videos obtained by various combinations of $c_{lb}$ and $c_{ub}$.
When $c_{lb}$ and $c_{ub}$ are set to 0.1 and  0.9 respectively, the majority of object appearances can be separated from scenes.


\begin{algorithm}[th]
\caption{Preprocessing algorithm.}
\label{alg:frame}
\textbf{Input}: Video frames $x_1^T$ \\
\textbf{Parameter}: $c_{lb}$, $c_{ub}$ and channel dimension $d_c$ \\
\textbf{Output}: Motion, scene and object videos

\begin{algorithmic}[1] 
\STATE Let $t=1$, $x_{s}=x_1$.
\WHILE{t $\leq$ T}
\STATE $x_{nxt}=x_T$ if ${t == T}$ else $x_{t+1}$
\STATE $m_t$ = $2 x_{t} - x_{s} - x_{nxt}$
\STATE $d_{pixel}$ = $max(abs(m_t), dim=d_c)$
\STATE $mask$ = $(d_{pixel} \geq c_{lb}) \odot (d_{pixel} \leq c_{ub})$
\STATE $o_t$ = $mask \odot x_t$
\STATE $s_t$ = $(1-mask) \odot x_t$
\ENDWHILE
\STATE \textbf{return} $m_1^T$, $s_1^T$ and $o_1^T$
\end{algorithmic}
\end{algorithm}

\begin{table*}[t]
    \centering
        \caption{Training settings of MOSO-VQVAE and MOSO-Transformer and quantitative results of video reconstruction on the UCF101, BAIR, KTH, RoboNet and KITTI datasets.}
        \label{tab:hyper}
        \scalebox{0.9}{
        \begin{tabularx}{1\linewidth}{ p{0.3\linewidth} | X<{\centering} X<{\centering} X<{\centering} X<{\centering} X<{\centering}  X<{\centering} X<{\centering}  X<{\centering}}
    \toprule
    Dataset     &  UCF101   &   BAIR    & \multicolumn{2}{c}{KTH}   &   \multicolumn{2}{c}{RoboNet} &   \multicolumn{2}{c}{KITTI}   \\
    Resolution  &   256     &   64      &   64      & 128           &   64      &   256             &   64  &   256     \\
    \midrule
    \textbf{MOSO-VQVAE} \\
    \midrule
    $T$         &   16      &   16      &   20      &   20          &   12      &   12              &   20  &   20      \\
    $f_{o}$    &   8       &   4       &   4       &   4           &   4       &   8               &   4   &   8       \\
    $f_{s}$    &   16      &   4       &   4       &   8           &   4       &   16              &   4   &   16      \\
    $f_{m}$    &   32      &   8       &   8       &   16          &   8       &   32              &   8   &   32      \\
    FPS         &   32      &   -       &   25      &   25          &   -       &   -               & -     & -   \\
    Batch size  &   2       &   24      &   16      &   6           &   24      &   3               &   24  &   2       \\
    Training steps& 250K    &   250K    &   250K    &   250K        &   250K    &  250K             &  300K &   300K     \\
    Learning rate&  2e-4    &   2e-4    &   2e-4    &   2e-4        &   2e-4    &   2e-4            &   2e-4&   2e-4    \\
    Scheduler   &  cosine   &   cosine  &   cosine  &   cosine      &   cosine  &   -               &   -   &   -       \\
    Discriminator Start Step &   50K     &   50K     &   50K     &   50K         &   50K     &   50K             &  50K  &   50K     \\
    \midrule
    PSNR        &   26.9    &   34.2    &   36.3    &   36.1        &   34.1    &   27.8            & 28.4  &  23.3     \\
    SSIM        &   75.7    &   95.9    &   94.4    &   93.2        &   94.8    &   83.4            & 87.8  &  63.8     \\
    LPIPS       &   0.190   &   0.010   &   0.030   &   0.044       &   0.013   &   0.072           & 0.042 &  0.241    \\
    \midrule
    \midrule
    \textbf{MOSO-Transformer}   \\
    \midrule
    Batch size  &   8       &   32      &   32      &   8           &   32      &   8               &   48  &   16       \\
    dropout     &   0.1     &   0.1     &   0.1     &  0.1          &   0.1     &   0.1             &  0.1  &   0.1     \\
Transformer$_{SO/G}$ Blocks&16& 16      &   16      &   16          &   16      &   16              &   7   &   12      \\
Transformer$_{M}$ Blocks& 8 &   8       &    8      &    8          &    8      &    8              &   7   &    7      \\
Attention heads &   8       &   8       &    8      &    8          &    8      &    8              &    8  &    8      \\
Embedding dim.  &   758     &   758     &   758     &  758          &   758     &   758             &  758  &   758     \\
Hidden dim.     &   1024    &   1024    &   1024    &  1024         &   1024    &   1024            &  1024 &  1024     \\
Immediate dim.  &   2048    &   2048    &   2048    &  2048         &   2048    &   2048            &  2048 &  2048     \\
Training steps  &   300K    &   90K     &   30K     &   85K         &   100K    &   110K            &  200K &  250K     \\
    \bottomrule
    \end{tabularx}}
\end{table*}
\begin{figure*}[t]
    \centering
    \includegraphics[width=1\linewidth]{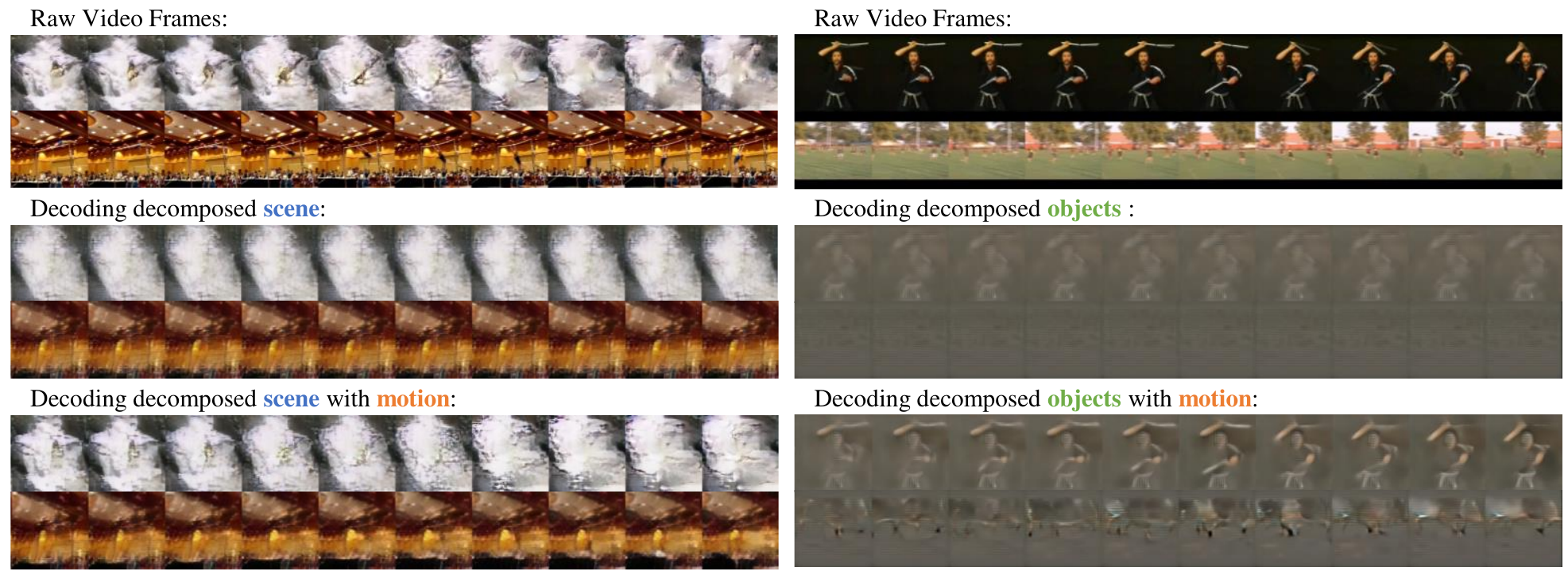}
    \caption{
    Visualizing decomposed objects and scenes with or without corresponding motions on UCF101.
    }
    \label{fig:decompose}
\end{figure*}

\begin{figure*}
    \centering
    \includegraphics[width=0.95\linewidth]{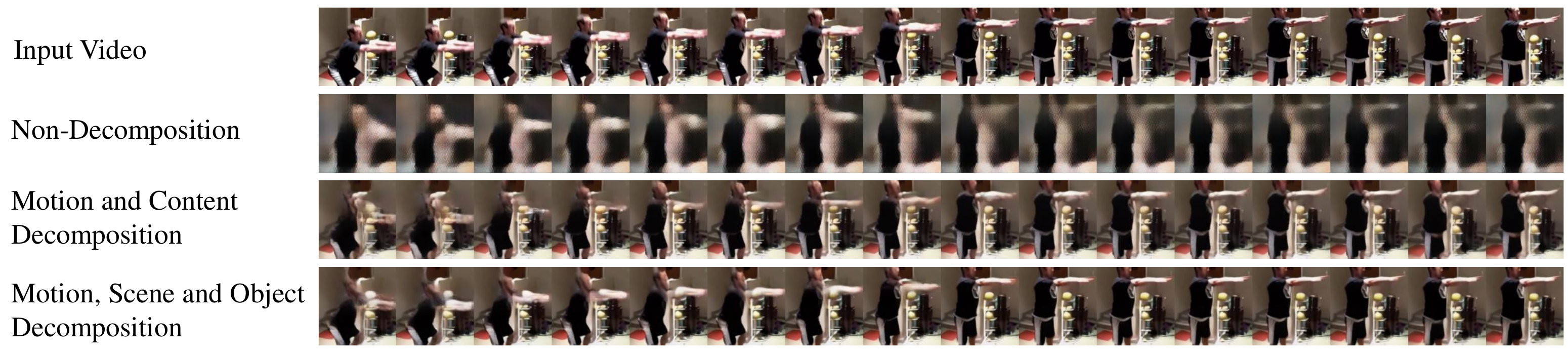}
    \caption{Qualitative comparison of ablated models on UCF101 for video reconstruction.
    The first row depicts the input video.
    The following three rows depict videos reconstructed by three ablated models.
    }
    \label{fig:ablate}
\end{figure*}

\section{A More General Situation of Eq. (12)}
When obtaining motion tokens, several downsample layers, i.e., 2D convolutions with stride 2, first downsample videos by frame.
The downsampled video frames are then concatenated in the temporal dimension and compose feature $z''_{mo} \in R^{H/f \times W/f \times T \times D}$.
The temporal self-attention splits the temporal dimension into $N_t$ working pools, obtains feature $z'_{mo} \in R^{H/f \times W/f \times N_t \times (T/N_t) \times D}$.
Each working pool contains features of $\frac{T}{N_t}$ consecutive video frames and \textbf{exchange of temporal information only happens between features in the same working pool}.
When $N_t=T$, no temporal information would be exchanged by the temporal self-attention, thus the $t$-th motion feature is obtained without the knowledge of other frames $x_{k}, k\neq t$.
Accordingly, any change of video frames $x_{k}, k\neq t$ will not affect the value of the $t$-th motion feature.
However, when $N_t=1$, video features are obtained by interacting between each pair of video frames, thus changes in any single video frame would affect values of all video features.
Considering the video prediction process of MOSO-Transformer, a pseudo video $\hat x_1^T$ is constructed through Eq. (10) and \textbf{has the same first K video frames} as the target video $x_1^T$.
By partitioning the first K video frames and the others into different working pools, the first K motion tokens of $x_1^K$ and $\hat x_1^K$ must be exactly \textbf{the same} as shown in Eq. (12).


A general solution to partition the given K given video frames and the subsequent ones to different working pools involves a constant hyper-parameter $c\in \{1,2,..,K\}$, which satisfies that K can be exactly divided by $c$.
Ensuring that $T$ can be exactly divided by $K$, then $N_t$ can be set as $\frac{cT}{K}$.
When $c=1$, the partition is the one stated in Eq. (12).
When $c=K$, then non-temporal information will be exchanged.

\section{Implementation Details and More Experimental Results}
\subsection{Hyperparameters and Optimizer}
MOSO is implemented with PyTorch \cite{pytorch}.
The specific training settings are given in Table \ref{tab:hyper}, where we denote the downsample factor in the motion, scene and object encoders as $f_{m}$, $f_{s}$ and $f_{o}$ respectively.
Adam optimizer \cite{adam} is used for both MOSO-VQVAE and MOSO-Transformer.

\subsection{Ablation Experimental Settings and Qualitative Results}
We conduct an ablation study to explore the necessity of decomposing object, scene and motion components.
Specifically, we compare the quality of decoded videos from (a) non-decomposed features, (b) content and motion decomposed features, and (c) scene, object and motion decomposed features.
To obtain non-decomposed video features, a frame-wise encoder is adopted to encode videos by frame and the merge module in the video decoder of MOSO-VQVAE is removed to reconstruct input videos.
The frame-wise encoder has the same settings and architecture as the motion encoder but takes raw video frames as input.
For content and motion decomposition encoding, a similar frame-wise encoder is used for encoding visual movements and a content encoder with the same structure as the scene encoder of MOSO-VQVAE is used to encode the content part.
The frame-wise encoder takes frame difference as input and the content encoder is fed with raw video frames.
The video decoder of MOSO-VQVAE is used to rebuild video details by summing frame-wise features with content features at multi scales.
For motion, scene and object decomposition, MOSO-VQVAE is used to encode decomposed video components and decode video details.
The total codebook size of all ablated models is 16384 and the dimension of all codebook entries is 256 for fair comparisons.
We visualize videos reconstructed by three ablated models in Fig. \ref{fig:ablate}, which demonstrates that our MOSO obtains more clear and more fidelity reconstruction results.

\begin{figure}
    \centering
    \includegraphics[width=0.95\linewidth]{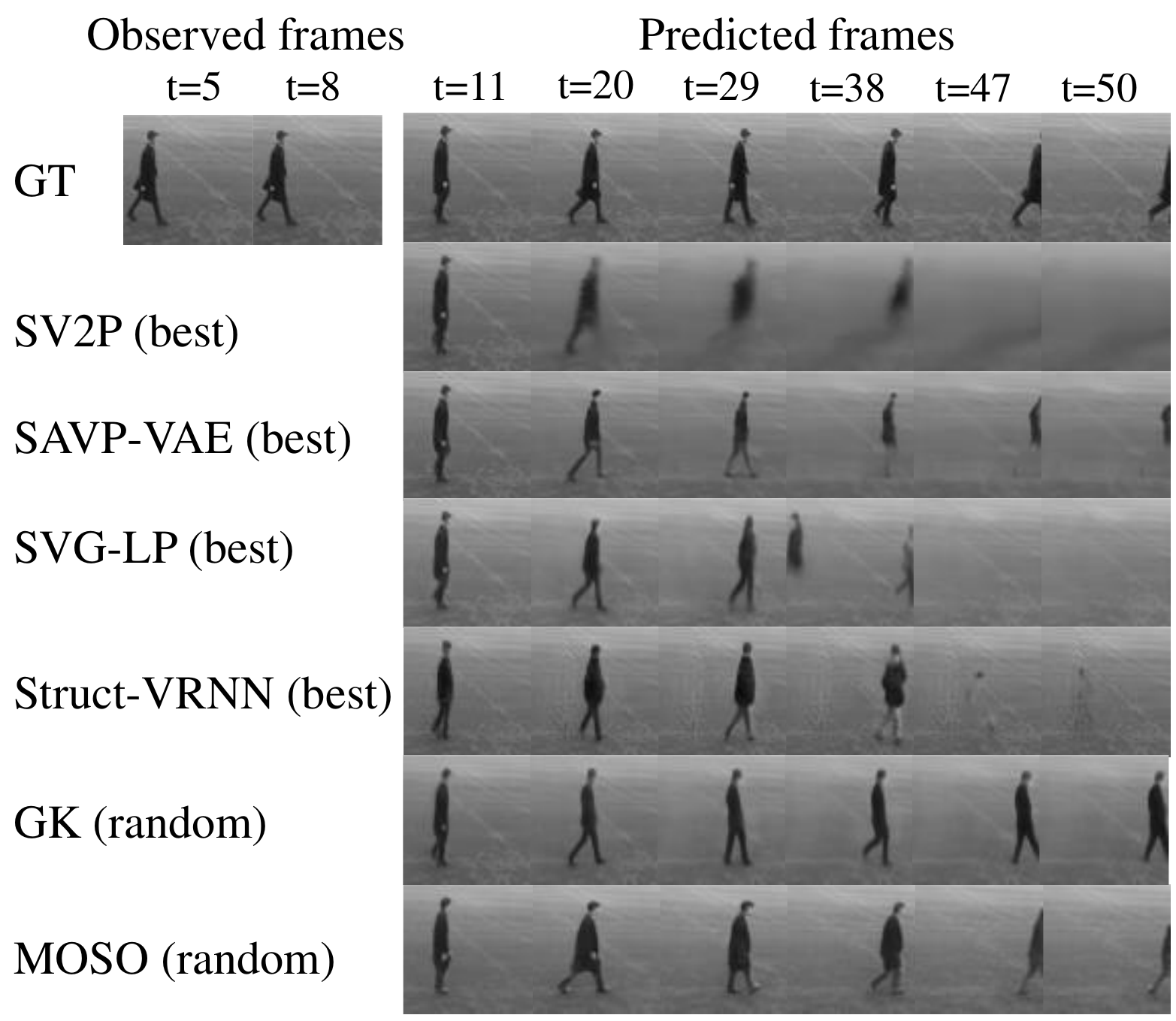}
    \caption{Qualitative comparison of MOSO and other models on KTH for video prediction.}
    \label{fig:kth}
\end{figure}

\subsection{Visualization of Decomposed Videos}
As stated in the paper, the encoded video features (i.e., motion, scene and object) can be decoded by the video decoder of MOSO-VQVAE flexibly.
Specifically, when decoding object features, we replace the scene and motion features with empty features filled with zeros and visualize the output of the video decoder.
When only replacing the scene features with empty features, we can decode objects with motion and observe corresponding motion patterns.
The decoding of scene features follows similar pipelines.
Samples of visualized components are given in Fig. \ref{fig:decompose}, which demonstrates that MOSO could well decompose scenes and objects and decouple different motion patterns.

\begin{figure}
\begin{minipage}[h]{0.48\linewidth}
    \centering
    \includegraphics[width=0.95\linewidth]{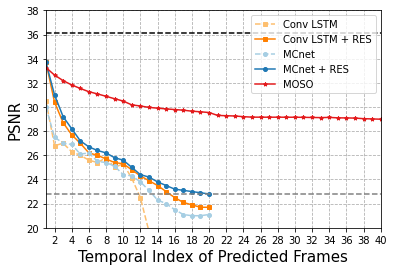}
\end{minipage}
\begin{minipage}[h]{0.48\linewidth}
    \centering
    \includegraphics[width=0.95\linewidth]{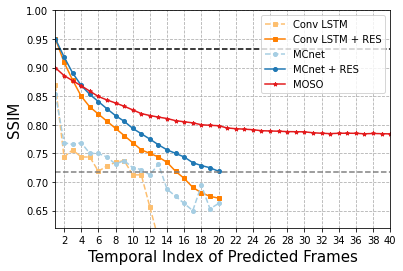}
\end{minipage}
\caption{
Quantitative comparison with prior work \cite{mocovp} on KTH $128^2$ for video prediction. 
The performance of MOSO declines more slowly over the temporal index of predicted video frames.
The black dashed line indicates the average reconstruction score.
}
\label{fig:kth128x}
\end{figure}

\subsection{More Ablation Studies}
\begin{table}[t]
    \centering
    \caption{
    Ablation study on video decomposition methods on KTH and UCF101 for video reconstruction. 
    \textit{non decom.}: non-decomposition;
    \textit{mo. co.}: motion and content decomposition;
    \textit{mo. sc. ob.}: motion, scene and object decomposition.
    \textit{pre. alg.} denotes the preprocessing algorithm.
    }
    \vspace{-1mm}
    \label{tab:decomp}
    \scalebox{0.8}{
    \begin{tabularx}{1.2\linewidth}{ p{0.23\linewidth} | X<{\centering} X<{\centering} p{0.1\linewidth} | X<{\centering} X<{\centering} p{0.1\linewidth}}
    \toprule
    \multirow{2}*{\textbf{Method}}  & \multicolumn{3}{c|}{KTH} & \multicolumn{3}{c}{UCF101} \\
                    &PSNR$\uparrow$ &SSIM$\uparrow$ &FVD$\downarrow$    &PSNR$\uparrow$ &SSIM$\uparrow$ & FVD$\downarrow$   \\
    \midrule
    non decom.      & 24.8          & 76.5          & 446.5             &   19.9        &   47.6        &  2487.2         \\
    \midrule
    mo. co.         & 32.6          & 86.4          & 238.8             &   28.5        &   75.8        &  1018.9       \\
    \midrule
    mo. sc. ob.     & 36.0          &\textbf{95.9}  & 237.8             &   29.8        &   79.6        &  310.1       \\
    + pre. alg.     & \textbf{36.5} &\textbf{95.9}  &\textbf{230.5}     & \textbf{30.0} & \textbf{80.6} &\textbf{267.9}\\
    \bottomrule
    \end{tabularx}
    }
\end{table}
\begin{table}[t]
    \centering
    \caption{
        Ablate discriminators in MOCO-VQVAE on UCF101. 
        \textit{$\mathcal{L}_{VD}$}: loss for video discriminator; 
        \textit{$\mathcal{L}_{ID}$}: loss for image discriminator;
        \textit{0.1/0.05}: loss weights.
    }
    \vspace{-1mm}
    \label{tab:disc}
    \scalebox{0.8}{
    \begin{tabularx}{1.2\linewidth}{ p{0.45\linewidth} | X<{\centering} X<{\centering} X<{\centering}}
    \toprule
    Methods             &SSIM$\uparrow$ &   LPIPS$\downarrow$&  FID$\downarrow$      \\
    \midrule
    w/o $\mathcal{L}_{VD}$/$\mathcal{L}_{ID}$           &  92.0     &   0.0294  &   24.8    \\
    \midrule
    0.1$\mathcal{L}_{VD}$               &   \textbf{92.1}    &   \textbf{0.0246}  &   \textbf{17.9}    \\
    0.1$\mathcal{L}_{VD}$ + 0.1$\mathcal{L}_{ID}$       &   90.7    &   0.0308  &   19.5   \\
    0.1$\mathcal{L}_{VD}$ + 0.05$\mathcal{L}_{ID}$      &   91.2    &   0.0277  &   19.7 \\
    \bottomrule
    \end{tabularx}
    }
    \vspace{-4mm}
\end{table}

\textbf{Video Decomposing}
We conduct an ablation study to explore the importance of motion, scene and object decomposition.
Specifically, we compare the quality of reconstructed videos of MOSO-VQVAE from (a) non-decomposed signals, (b) content and motion decomposed signals, and (c) scene, object and motion decomposed signals on two benchmarks, i.e., KTH and UCF101.
Settings for each ablated model are given in the appendix and the results are given in Table \ref{tab:decomp}.
By separating content from motion, the quality of rebuilt videos improves on all metrics and benchmarks.
When further separating objects from scenes, the reconstruction quality further enhances.
After adopting our simple but effective preprocess algorithm, our MOSO-VQVAE achieves the best reconstruction quality on both the UCF101 and KTH datasets.

\textbf{Adversarial Training}
Inspired by VQGAN \cite{taming}, we adopt video and/or image discriminators to train MOSO-VQVAE in an adversarial manner.
The video and image discriminators respectively evaluate videos by clip and by frame.
As shown in Table \ref{tab:disc}, using a video discriminator with a loss weight of 0.1 achieves the best LPIPS and FID and comparable SSIM.
The image discriminator brings no improvement since it cannot preserve video consistency when optimizing a reconstructed video frame.

\textbf{Codebook Sharing}
We conduct an ablation study on the shared codebook as reported in Table \ref{tab:cb}.
When sharing codebooks to quantize different features (i.e. motion, scene and object features), we obtain better reconstruction performance with a smaller total codebook size.
There exist two potential causes.
Firstly, similarly to quantizing features with multi-scales, 
which has improved the performance of VQ-VAE on image reconstruction \cite{lverse},
quantizing features with multi-perspectives can make the codebook more diverse and informative.
Second, the regions of features obtained by the three encoders may partially but not entirely overlap.
Thus codebook sharing boosts performance with a one-third reduction in total codebook size, e.g., 16384 versus $8192\times3$.
In contrast, when the shared codebook is the same size as each individual codebook, performance degrades due to codebook sharing.
\begin{table}[t]
    \centering
    \caption{
        Ablate codebooks in MOCO-VQVAE on UCF101. 
        \textit{sep. cb.}: each encoder adopts an independent codebook; 
        \textit{share cb.}: sharing codebooks used for three encoders;
        \textit{$N$}: the codebook size.
    }
    \vspace{-3mm}
    \label{tab:cb}
    \scalebox{0.8}{
    \begin{tabularx}{1.2\linewidth}{ p{0.2\linewidth} | p{0.25\linewidth} | X<{\centering} X<{\centering} X<{\centering}}
    \toprule
    Methods                 & $N$               &SSIM$\uparrow$ &   LPIPS$\downarrow$&  FID$\downarrow$      \\
    \midrule
    sep. cb.                &   8192 $\times$ 3 & 88.5          &  0.0306           &  17.5     \\
    \midrule
    \multirow{2}*{share cb.}&   8192            & 88.0          &  0.0335           &  17.9    \\
                            &   16384           & \textbf{89.4} &  \textbf{0.0294}  &  \textbf{16.5}   \\
    \bottomrule
    \end{tabularx}
    }
\end{table}

\subsection{More Experimental Results}
We qualitatively compare MOSO with prior works on KTH at $64^2$ resolution in Fig. \ref{fig:kth}.
When SV2P \cite{sv2p}, SAVP-VAE \cite{savp}, SVG-LP \cite{svg-lp} and Struct-VRNN \cite{vrnn} fail to synthesize consistent human objects in the last several frames, GK \cite{grid} and our MOSO could predict a long future video with consistent object identities and reasonable subsequent actions.
Moreover, our MOSO generates more distinct object identities and more realistic actions.
The better performance benefits from the decomposition of motion, scene and object, which helps to model varied motions and reduce disturbance of motion artifacts on object identities.

At $128^2$ resolution, we compare MOSO with prior work \cite{mocovp} quantitatively in Fig. \ref{fig:kth128x}.
The black dashed line represents the average reconstruction score of MOSO-VQVAE on PSNR and SSIM, which becomes the upper bound for MOSO-Transformer on video prediction.
We train MOSO-VQVAE with negative SSIM loss on all videos and remove the discriminator loss on KTH $128^2$.
As shown in Fig. \ref{fig:kth128x}, MOSO outperforms prior work \cite{mocovp} on PSNR and SSIM after the 2nd and 5th predicted frames respectively, and the performance of MOSO declines much more slowly over time, demonstrating its potential on generating long videos.

\subsection{Additional Samples}
To facilitate visualization, we provide additional samples of MOSO via a website: \hyperlink{https://iva-mzsun.github.io/MOSO}{https://iva-mzsun.github.io/MOSO}.

{\small
\bibliographystyle{ieee_fullname}
\bibliography{egbib}
}

\end{document}